\documentclass{article}

\usepackage[preprint]{neurips_2026}

\usepackage[utf8]{inputenc}
\usepackage[T1]{fontenc}
\usepackage{hyperref}
\usepackage{url}
\usepackage{booktabs}
\usepackage{amsfonts}
\usepackage{nicefrac}
\usepackage{microtype}
\usepackage{xcolor}
\usepackage{textgreek}
\usepackage{amsmath}
\usepackage{amssymb}
\usepackage{graphicx}
\usepackage{subcaption}
\usepackage{multirow}
\usepackage{array}
\usepackage{bbm}

\title{Cardinality-Decomposed Loss: Matching Training Objectives to Relation Structure in Heterogeneous Recommendation Graphs}

\author{%
  {\bf Parul Maheshwari, Amulya Paruchuri, Yiqing Zou,} \\
  {\bf Alireza Sahami Shirazi, Farhad Farahani, Prakhar Mehrotra} \\
  PayPal AI \\
  PayPal Inc. \\
  USA
}

\begin{document}
\maketitle

\begin{abstract}
Graph Neural Networks trained on heterogenous bipartite graphs form a common basis in recommendation systems. These graphs often express relations that vary in cardinality, for example, user-item preferences are one-to-many and user-attribute features are one-to-one. Traditionally, a unique loss function is applied for all of the network components which is often Bayesian Personalized Ranking (BPR). While BPR works well for the recommendation task, we find that it causes attribute embeddings to collapse to near-random geometry --- a silent failure that leaves standard ranking metrics largely unaffected and therefore invisible to conventional evaluation. This in turn pollutes user node embeddings, which are shaped by both edge types simultaneously, hurting downstream tasks like personalization, segmentation, etc. Here we propose a Cardinality-Decomposed Loss (CDL) that combines both Cross Entropy (CE) and BPR to enable the model to collectively optimize for relations across cardinalities. As we implement this loss, we also confirm the conflict between CE and BPR by showing that the two losses compete against each other in the shared encoder's parameter space. We evaluate CDL on five datasets spanning two structural configurations --- one-to-one attributes on user nodes (MovieLens-1M, Last.fm-360K, PayPal Audience Factory, BookCrossing) and on item nodes (Yelp) --- and find that CDL consistently improves discriminability in attribute embeddings. We also show that whenever these attributes contain meaningful preference signal, we also see improvement in the ranking task (measured by NDCG). On the other hand, when attributes are weakly correlated with preferences, there is an inherent tension between the two objectives. We use a lambda parameter to navigate this trade-off, and a lambda-sweep reveals that dataset behavior is governed by two graph properties --- semantic alignment and topology leakage. Semantic alignment captures whether the one-to-one attribute is predictive of user preferences, while topology leakage captures whether message passing already encodes attribute structure implicitly through the graph's connectivity.
\end{abstract}

\section{Introduction}
\label{sec:intro}

Graph Neural Networks (GNNs) have become the dominant backbone for
collaborative-filtering recommendation systems~\cite{ngcf,lightgcn,pinsage}.
In practice, the input graphs are \emph{heterogeneous}: alongside the primary
user--item interaction edges, practitioners encode rich auxiliary structure such
as user demographics, item genres, geographic attributes, and behavioural
segments. These different relation types have fundamentally different
\emph{structural cardinalities}. A user--item edge is \emph{one-to-many}: a
single user rates many items and there is no exhaustive enumeration of the
``non-rated'' items. By contrast, a user--age edge is \emph{one-to-one}: the
set of possible ages is small and fully enumerated, and each user is assigned
exactly one.

Despite this structural difference, nearly all heterogeneous GNN pipelines apply
the same loss to both relation types: Bayesian Personalized Ranking
(BPR)~\cite{bpr}. BPR is derived under a generative model that assumes the set
of unobserved interactions is a representative sample of truly unpreferred items
--- an assumption that is satisfied for sparse preference data but \emph{violated}
for exhaustive discrete attributes. Applying BPR to a user--gender edge means
treating ``not being female'' as evidence of dispreference for the male node, and
sampling random gender nodes as negatives. This is semantically incoherent.

We show that this mismatch produces a \textbf{silent representational failure}.
Attribute embeddings (gender, age, occupation, city, country) collapse toward
near-random geometry under uniform BPR training: in our experiments, the
mean cosine similarity between distinct attribute node embeddings reaches 0.93--0.98
after 50 epochs. Crucially, this collapse is \emph{invisible to standard ranking
metrics} --- NDCG@10 and Hit@10 remain largely unaffected because the ranking task
never directly evaluates attribute embedding quality. Practitioners who rely on
these metrics alone will not detect the failure.

This matters for any downstream use of node embeddings: personalization, user
segmentation, cold-start initialisation, and fairness auditing all depend on
attribute structure being geometrically recoverable from the learned
representations. Silent collapse contaminates not just attribute embeddings but
\emph{user embeddings} as well, since both are shaped by the same shared message-passing
encoder.

This paper identifies and characterises the silent failure (\S\ref{sec:failure}),
then proposes \textbf{Cardinality-Decomposed Loss (CDL)} as the fix: a 
modification that assigns training objectives by structural cardinality --- BPR for
one-to-many relations, cross-entropy for one-to-one --- combined as
$\mathcal{L}_{\mathrm{CDL}} = \mathcal{L}_{\mathrm{BPR}} + \lambda \cdot \mathcal{L}_{\mathrm{CE}}$
(\S\ref{sec:method}). We confirm the structural mismatch mechanistically by showing
that gradient cosine similarity between the two loss terms is consistently negative
in the shared encoder, providing direct evidence that BPR and CE compete on the
parameters they share (\S\ref{sec:gradients}). Across five datasets, CDL consistently
recovers attribute discriminability (+30--42~pp in linear probe AUC) and improves
ranking quality where one-to-one attributes are semantically predictive of preference:
+7.8\% NDCG@10 on Last.fm-360K, +2.9\% on Yelp, and +3.3\% on a large-scale
industrial graph with 1M users and 178M interactions (\S\ref{sec:experiments}).
Finally, we introduce a two-axis framework of \emph{semantic alignment} and
\emph{topology leakage} that characterises the $\lambda$-sweep behaviour across
datasets and allows practitioners to predict which regime they will encounter
before running experiments (\S\ref{sec:analysis}).

\section{Related Work}
\label{sec:related}

Graph-based collaborative filtering has converged on BPR as its standard training objective. NGCF~\cite{ngcf} and LightGCN~\cite{lightgcn} established this pattern on user--item bipartite graphs, and PinSage~\cite{pinsage} carried it to web scale via GraphSAGE~\cite{graphsage}. Because these systems operate on homogeneous graphs with a single edge type, the structural appropriateness of BPR is never questioned.  Extending GNNs to heterogeneous graphs --- as in RGCN~\cite{rgcn}and HAN~\cite{han} introduces relation-specific message passing, but the training objective is typically not revisited. Knowledge-graph-enhanced recommenders such as KGCN~\cite{kgcn} and RippleNet~\cite{ripplenet}incorporate item-side attribute edges yet apply BPR uniformly or add manually-tuned auxiliary objectives without principled justification. CDL provides that basis: relation cardinality, readable directly from the data schema, determines which objective is structurally correct.

BCIPM~\cite{bcipm}is the closest prior work, combining BPR and BCE via $\mathcal{L} = \beta \cdot \mathcal{L}_{\text{bce}} + (1-\beta) \cdot \mathcal{L}_{\text{bpr}}$. but assigns losses by model component rather than relation structure, and collapses all relation types into a homogeneous graph during pre-training. CDL instead derives the BPR-versus-BCE assignment from structural cardinality, preserving heterogeneous graph structure throughout training.
The broader multi-task learning literature addresses gradient conflict between
objectives~\cite{graddrop,cagrad} and the combination of related prediction tasks
in recommendation~\cite{mmoe,ple}. Our setting is distinct: the conflict arises between two loss formulations applied to structurally different edge types within a single model, and we eliminate its root cause by using the correct loss per relation type rather than resolving conflict post hoc. Prior work noting geometric embedding degeneracy under ranking-focused training~\cite{anisotropy,uniformity}treats this as a general property of the training signal; we identify the more specific cause --- cardinality mismatch in heterogeneous graphs --- and provide a targeted fix requiring no changes to model architecture.

\section{Problem Setup}
\label{sec:setup}

\paragraph{Heterogeneous bipartite graph.}
Let $\mathcal{G} = (\mathcal{V}, \mathcal{E}, \phi, \psi)$ be a heterogeneous graph
with node type mapping $\phi: \mathcal{V} \to \mathcal{T}$ and edge type mapping
$\psi: \mathcal{E} \to \mathcal{R}$. We focus on graphs with a user node set
$\mathcal{U}$ and multiple item/attribute node sets. The edge set decomposes as
$\mathcal{E} = \bigcup_{r \in \mathcal{R}} \mathcal{E}_r$.

\paragraph{Relation cardinality.}
We distinguish two structural cardinality types:
\begin{itemize}
\item \textbf{One-to-many (OTM)}: A source node $u$ can be connected to a variable
  number of destination nodes $\{v_1, v_2, \ldots\}$, drawn from a large item space.
  The unobserved edges are a representative sample of unpreferred items. Example:
  user--item interaction edges.

\item \textbf{One-to-one (OTO)}: Each source node $u$ is connected to \emph{exactly one}
  destination node $v \in \{1, \ldots, K\}$ from a small fully-enumerated set of $K$
  classes. The node assignment is exhaustive: $u$ is assigned to one class and none of
  the remaining $K-1$ classes. Example: user--gender, user--age-group, item--city.
\end{itemize}

\paragraph{Training objectives.}
For OTM relations, BPR~\cite{bpr} is the standard choice:
\begin{equation}
  \mathcal{L}_{\mathrm{BPR}} = -\mathbb{E}_{(u,v^+,v^-)} \left[ \log \sigma\!\left(
  \mathbf{e}_u^\top \mathbf{e}_{v^+} - \mathbf{e}_u^\top \mathbf{e}_{v^-} \right)\right],
  \label{eq:bpr}
\end{equation}
where $v^-$ is sampled uniformly from items not interacted with by $u$. For OTO
relations, cross-entropy (CE) is the structurally correct choice:
\begin{equation}
  \mathcal{L}_{\mathrm{CE}} = -\mathbb{E}_u \left[ \log \frac{\exp(\mathbf{w}_{y_u}^\top
  \mathbf{e}_u)}{\sum_{k=1}^K \exp(\mathbf{w}_k^\top \mathbf{e}_u)} \right],
  \label{eq:ce}
\end{equation}
where $y_u \in \{1,\ldots,K\}$ is the ground-truth class label and $\mathbf{w}_k$ are
learned classifier weights. For nodes missing an attribute label, CE supervision is
masked and those nodes still participate in BPR training.

\section{The Silent Failure: BPR on One-to-One Relations}
\label{sec:failure}

We first establish the failure mode that motivates CDL.
\paragraph{BPR is generatively wrong for OTO relations.}
BPR assumes a generative model $P(u \text{ prefers } v^+ \text{ over } v^-) \propto
e^{\mathbf{e}_u^\top \mathbf{e}_{v^+} - \mathbf{e}_u^\top \mathbf{e}_{v^-}}$, which is
meaningful for preference data where $v^-$ is an item the user might conceivably prefer.
For a gender edge, the ``negative'' gender node is the one to which the user is
definitively \emph{not} assigned. Sampling it as a ``dispreferred'' item conflates
structural exclusion with preference, forcing the model to push gender node embeddings
apart in a ranking sense rather than organising them as discriminative class representatives.

\paragraph{Embedding collapse.}
We train a baseline model using BPR on all relation types (the ``\texttt{uniform}'' mode)
and measure attribute embedding geometry after 50 epochs. Mean cosine similarity between
distinct attribute node pairs reaches 0.980 (MovieLens-1M), 0.958 (BookCrossing), and
0.934 (Audience Factory) --- in all cases 100\% of pairs exceed the conservative 0.5
collapse threshold. A random embedding initialised uniformly on the unit sphere has
expected cosine similarity $\approx 0$; values of 0.93--0.98 indicate that the model
has learned to ignore attribute structure entirely.

\begin{figure*}[t]
\centering
\includegraphics[width=\textwidth]{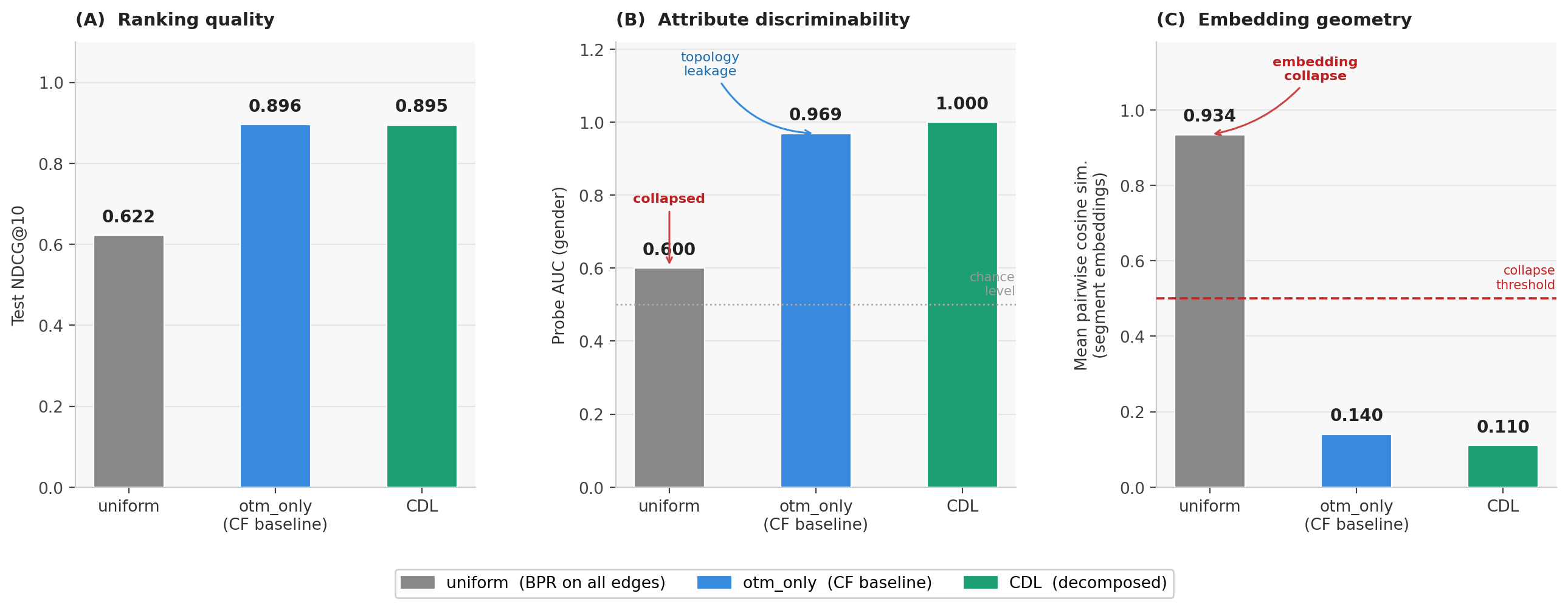}
\caption{\textbf{The silent failure on PayPal Audience Factory.} All three panels share
the same x-axis (training mode: uniform BPR, CF baseline, CDL).
\textbf{(A)} NDCG@10 correctly distinguishes uniform BPR (0.622) as weaker, but cannot
distinguish the CF baseline from CDL (0.896 vs.\ 0.895) --- the metric is blind to the
representational failures panels (B) and (C) expose.
\textbf{(B)} Linear probe AUC on gender embeddings reveals the divergence. Uniform BPR
collapses attribute representations to near-chance discriminability (0.600); the CF
baseline reaches 0.969 through topology leakage, since gender nodes are directly adjacent
in the graph and message passing propagates demographic structure without any explicit
supervision; CDL recovers perfect discriminability (1.000).
\textbf{(C)} Mean pairwise cosine similarity between segment embeddings confirms geometric
collapse under uniform BPR (0.934 --- well above the 0.5 collapse threshold), while both
the CF baseline and CDL maintain healthy embedding diversity.}
\label{fig:silent_failure}
\end{figure*}

\paragraph{Invisibility to ranking metrics.}
Critically, this collapse has a small effect on NDCG@10 in some datasets. For MovieLens-1M,
the \texttt{uniform} baseline achieves NDCG@10 of 0.296 at 50 epochs --- comparable to
the CF baseline's 0.399 at 150 epochs when both are learning. The collapse is not caught
by the metric used to monitor training. We term this a \emph{silent} failure: the model
appears healthy by the metrics practitioners inspect, while its representations are
structurally broken. Figure~\ref{fig:silent_failure} illustrates all three dimensions
of the failure simultaneously on the Audience Factory dataset.

\section{Cardinality-Decomposed Loss}
\label{sec:method}

\begin{figure*}[t]
\centering
\includegraphics[width=\textwidth]{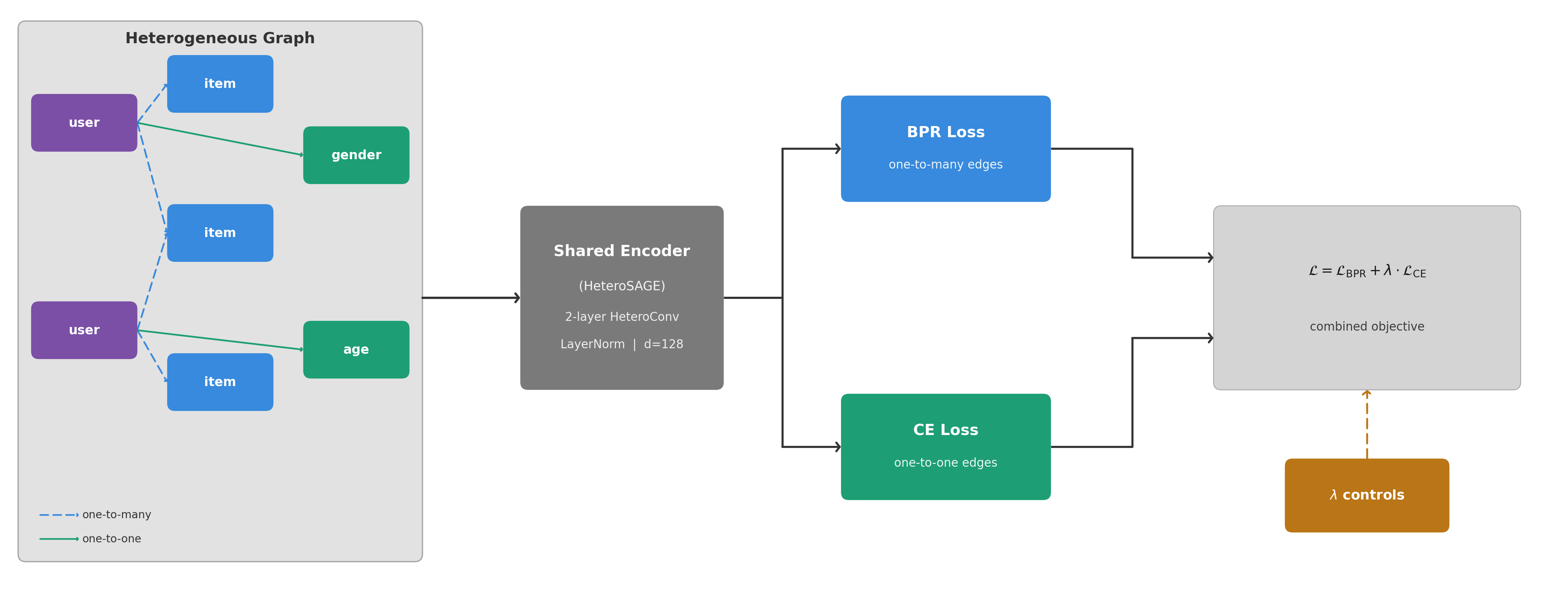}
\caption{\textbf{CDL architecture overview.} \textbf{Left:} The heterogeneous input graph
contains two structurally distinct edge types: one-to-many user--item edges (dashed blue)
and one-to-one user--attribute edges (solid teal). \textbf{Centre:} Both edge types pass
through a shared two-layer HeteroSAGE encoder, allowing attribute structure to inform
user representations via message passing. \textbf{Right:} The encoder output feeds two
separate loss heads --- BPR for one-to-many edges and cross-entropy classification heads
for one-to-one edges --- combined as $\mathcal{L} = \mathcal{L}_{\mathrm{BPR}} + \lambda
\cdot \mathcal{L}_{\mathrm{CE}}$. The scalar $\lambda$ is the only new hyperparameter;
it is not learned but swept over a small grid.}
\label{fig:architecture}
\end{figure*}

\paragraph{CDL formulation.}
Given a cardinality map $\kappa: \mathcal{R} \to \{\mathrm{OTM}, \mathrm{OTO}\}$ assigned
from structural inspection of the graph (Section~\ref{sec:setup}), CDL combines
objectives as:
\begin{equation}
  \mathcal{L}_{\mathrm{CDL}} = \mathcal{L}_{\mathrm{BPR}} + \lambda \cdot
  \mathcal{L}_{\mathrm{CE}},
  \label{eq:cdl}
\end{equation}
where $\mathcal{L}_{\mathrm{BPR}}$ aggregates over all OTM relations and
$\mathcal{L}_{\mathrm{CE}}$ aggregates over all OTO relations. The scalar
$\lambda \geq 0$ controls the relative weight and is the only hyperparameter
introduced beyond the base GNN configuration.

For each OTO relation $r$ with $K_r$ classes, we attach a dedicated linear
classification head $f_r: \mathbb{R}^d \to \mathbb{R}^{K_r}$. These heads are
trained jointly with the GNN backbone. CE loss is computed per attribute and averaged.
Missing labels are handled via a per-attribute mask tensor
$\mathbf{m}_r \in \{0,1\}^{|\mathcal{U}|}$; CE is summed only over nodes with
$m_{r,u} = 1$.

The cardinality map $\kappa$ is determined by inspecting the graph's adjacency statistics:
a relation is OTO if (i) the source-to-target degree is exactly one for all nodes with
the relation and (ii) the target node set is fully enumerated with $K \ll |\mathcal{U}|$.
Both conditions are verifiable from the data schema before any training is run.

\paragraph{Baselines.}
We compare CDL against \texttt{otm\_only} (BPR on OTM edges only; OTO edges present for message passing but not supervised --- the fair CF baseline) and \texttt{uniform} (BPR on all relation types --- the broken baseline confirming the failure mode, excluded from primary comparison as structurally unsound).

\section{Mechanistic Evidence: Gradient Conflict}
\label{sec:gradients}

CDL's combination of BPR and CE on a shared encoder raises the question: do the two
losses cooperate or compete? We answer this by measuring the gradient cosine
similarity between the two loss terms at each training step.

\paragraph{Measurement protocol.}
Let $\Theta_{\mathrm{shared}}$ denote the parameters of the shared encoder that
touch the node type carrying OTO attributes (user parameters for user-side OTO
relations, item parameters for item-side OTO). Let
$\mathbf{g}_{\mathrm{BPR}} = \nabla_{\Theta_{\mathrm{shared}}} \mathcal{L}_{\mathrm{BPR}}$
and $\mathbf{g}_{\mathrm{CE}} = \nabla_{\Theta_{\mathrm{shared}}} \mathcal{L}_{\mathrm{CE}}$.
We log:
\begin{equation}
  \rho = \cos(\mathbf{g}_{\mathrm{BPR}},\, \mathbf{g}_{\mathrm{CE}}) =
  \frac{\mathbf{g}_{\mathrm{BPR}}^\top \mathbf{g}_{\mathrm{CE}}}
       {\|\mathbf{g}_{\mathrm{BPR}}\|\,\|\mathbf{g}_{\mathrm{CE}}\|}.
\end{equation}
A value $\rho < 0$ indicates that improving the BPR objective hurts the CE objective
on the shared parameters, and vice versa.

Table~\ref{tab:gradients} reports gradient cosine similarity measurements across
datasets. Conflict is present from the first training step in all settings.

\begin{table}[h]
\centering
\caption{Gradient cosine similarity $\rho$ between $\mathcal{L}_{\mathrm{BPR}}$
and $\mathcal{L}_{\mathrm{CE}}$ in the CDL model. Negative values indicate
genuine conflict on shared encoder parameters. Measurements are at step 0 (first
gradient) for Last.fm and throughout 150-epoch training for Audience Factory.}
\label{tab:gradients}
\vspace{2pt}
\begin{tabular}{lccc}
\toprule
Dataset & $\rho$ at step 0 & $\rho$ mean (150 ep.) & $\rho$ min (150 ep.) \\
\midrule
MovieLens-1M  & $+0.032$ & $-0.033$ & $-0.307$ \\
Last.fm-360K  & $-0.097$ & --- & --- \\
Yelp          & --- & $+0.0003$ & $-0.271$ \\
BookCrossing  & --- & $0.000$ & $-0.291$ \\
Audience Factory & $-0.129$ & $-0.044$ & $-0.307$ \\
\bottomrule
\end{tabular}
\end{table}

We observe two qualitatively distinct patterns. In \emph{Pattern~A} (MovieLens), conflict begins near-zero in epoch~1 and transitions to consistently negative as both loss surfaces develop strong gradients, peaking around epoch~10--20 before partial recovery. In \emph{Pattern~B} (Last.fm, Audience Factory), conflict is \emph{negative from epoch~1} because the demographic distribution is strongly orthogonal to the preference signal --- on Last.fm-360K, 73\% of users are male yet music taste is not aligned with this demographic skew. Both patterns confirm that BPR and CE objectives compete on shared encoder parameters; Pattern~B datasets require larger $\lambda$ to let CE reshape the embedding space against stronger BPR dominance.

\section{Experiments}
\label{sec:experiments}

\subsection{Datasets}

We evaluate on five datasets spanning diverse domains and structural configurations
(Table~\ref{tab:datasets}). Four datasets have OTO attributes on user nodes
(MovieLens-1M, Last.fm-360K, Audience Factory, BookCrossing); one has OTO attributes
on item nodes (Yelp).

\begin{table}[h]
\centering
\caption{Dataset statistics after preprocessing and interaction filtering.}
\label{tab:datasets}
\vspace{2pt}
\resizebox{\columnwidth}{!}{%
\begin{tabular}{lrrrlll}
\toprule
Dataset & Users & Items & Interactions & OTO attributes & OTO node & Split \\
\midrule
MovieLens-1M & 6,040 & 3,706 & 988K & gender (2), age (7), occ.\ (21) & user & LOO \\
Last.fm-360K & 263,671 & 145,431 & 12.3M & gender (2), age (6), country (21) & user & LOO \\
Yelp & $\sim$200K & $\sim$60K & 2.7M & city (31), state (15), stars (5) & business & LOO \\
BookCrossing & 15,223 & 36,041 & 534K & age (6, 68\% coverage) & user & LOO \\
Audience Factory & 1M & 242 & 6.2M & gender (2), age (3, 83\% cov.) & user & LOO \\
\bottomrule
\end{tabular}}
\end{table}

\paragraph{Preprocessing.}
All datasets use leave-one-out (LOO) splits: for each user, the last item is held
out for test and the second-to-last for validation, with the remainder used for
training. Negatives for evaluation are pre-sampled at 99 per user and fixed across
all runs. For Last.fm-360K, users with fewer than 20 interactions or incomplete
demographic profiles are excluded. For Audience Factory, interactions are deduplicated
to unique (user, segment) pairs; users with fewer than three distinct segment dates
are train-only.

\subsection{Architecture and Training}

\paragraph{Model.}
We use a two-layer \textbf{HeteroSAGE} for all experiments: a \texttt{HeteroConv}
wrapper applying \texttt{SAGEConv}$(d, d, \text{normalize=True})$ per edge type,
followed by \texttt{LayerNorm} per node type, with dropout 0.1 between layers.
User nodes are initialised with a learnable embedding of dimension $d=128$. All
other node types are initialised with 384-dimensional sentence embeddings from
\texttt{all-MiniLM-L6-v2}~\cite{sentence-bert}, projected to $d=128$ by a linear
layer before the first GNN layer. Sentence-transformer weights are frozen after
feature extraction.

\paragraph{Training.}
All models are trained with Adam ($\text{lr}=10^{-3}$, weight\_decay $= 10^{-5}$),
early stopping with patience 10 on validation NDCG@10, and a 150-epoch budget.
The training loop uses a full-graph forward pass to obtain all node embeddings,
followed by mini-batch BPR and CE loss computation over pre-sampled triplets.
All experiments were run on a single H100 GPU (80~GB HBM); the 150-epoch runs
completed in under 12 hours per dataset on the four public benchmarks.

\subsection{Evaluation Metrics}

\textbf{Ranking:} NDCG@10 and Hit@10 in the 100-item retrieval setting (1 positive
+ 99 fixed negatives). \textbf{Attribute discriminability:} Linear probe AUC ---
a logistic regression classifier trained on frozen user embeddings evaluated on
held-out users; macro-AUC per attribute. \textbf{Neighborhood purity:} $k$-NN
purity with $k=10$ in cosine space --- the fraction of a node's 10 nearest
neighbors that share the same attribute class.

\subsection{Main Results}
\label{subsec:main}

Table~\ref{tab:main} reports the primary comparison between CDL (\texttt{decomposed})
and the CF baseline (\texttt{otm\_only}) at the optimal $\lambda$ for each dataset.

\begin{table*}[t]
\centering
\caption{Main results at 150-epoch budget. ``CDL (best $\lambda$)'' uses the
$\lambda$ that maximises test NDCG@10 (see Table~\ref{tab:lambda} for the full
sweep). Attribute AUC is macro-averaged over all OTO attributes for that dataset.
Bold indicates best result per metric. $\Delta$NDCG is the relative change
from the CF baseline.}
\label{tab:main}
\vspace{2pt}
\resizebox{\textwidth}{!}{%
\begin{tabular}{llcccccc}
\toprule
Dataset & Mode & $\lambda$ & Test NDCG@10 & Test Hit@10 & $\Delta$NDCG & Probe AUC (macro) & kNN Purity (macro) \\
\midrule
\multirow{2}{*}{MovieLens-1M}
  & \texttt{otm\_only} (CF baseline) & --- & 0.399 & 0.650 & --- & 0.700 & 0.394 \\
  & \texttt{CDL} & 1.0 & 0.336 & 0.580 & $-15.8\%$ & \textbf{0.837} & \textbf{0.987} \\
\midrule
\multirow{2}{*}{Last.fm-360K}
  & \texttt{otm\_only} (CF baseline) & --- & 0.592 & 0.866 & --- & 0.664 & 0.363 \\
  & \texttt{CDL} & 2.0 & \textbf{0.638} & \textbf{0.871} & $\mathbf{+7.8\%}$ & \textbf{0.989} & \textbf{0.465} \\
\midrule
\multirow{2}{*}{Yelp}
  & \texttt{otm\_only} (CF baseline) & --- & 0.589 & 0.822 & --- & 0.620 & 0.721 \\
  & \texttt{CDL} & 1.0 & \textbf{0.606} & \textbf{0.840} & $\mathbf{+2.9\%}$ & \textbf{0.766} & \textbf{1.000} \\
\midrule
\multirow{2}{*}{BookCrossing}
  & \texttt{otm\_only} (CF baseline) & --- & \textbf{0.304} & \textbf{0.516} & --- & 0.642 & 0.293 \\
  & \texttt{CDL} & 2.0 & 0.282 & 0.481 & $-7.2\%$ & \textbf{0.943} & \textbf{1.000} \\
\midrule
\multirow{2}{*}{Audience Factory}
  & \texttt{otm\_only} (CF baseline) & --- & 0.422 & 0.663 & --- & 1.000\textsuperscript{$\dagger$} & 0.774 \\
  & \texttt{CDL} & 1.0 & \textbf{0.435} & \textbf{0.677} & $\mathbf{+3.3\%}$ & \textbf{1.000} & \textbf{1.000} \\
\bottomrule
\end{tabular}}
\vspace{2pt}
\footnotesize $^\dagger$ Probe AUC = 1.0 in the CF baseline due to topology leakage --- gender/age nodes are connected
to users in the graph; message passing already encodes demographic structure (see \S\ref{sec:analysis}).
\end{table*}

\begin{figure*}[t]
\centering
\includegraphics[width=\textwidth]{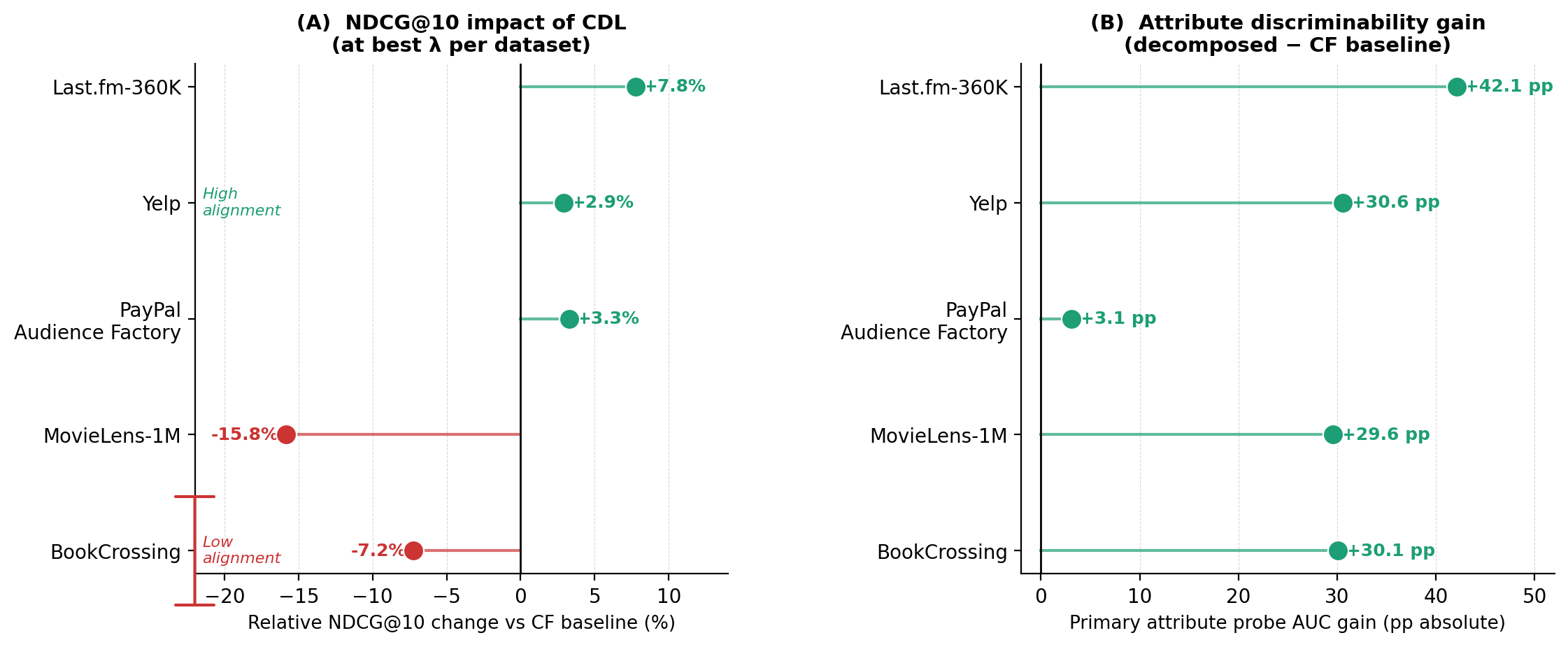}
\caption{\textbf{CDL impact across all five datasets.} \textbf{Left:} NDCG@10 change
relative to the CF baseline at the best $\lambda$ per dataset (positive = CDL improves
ranking). Datasets with high semantic alignment (Last.fm $+7.8\%$, Audience Factory
$+3.3\%$, Yelp $+2.9\%$) sit above zero; low-alignment datasets (MovieLens, BookCrossing)
incur a controlled ranking cost. The vertical dashed line marks zero. \textbf{Right:}
Absolute gain in primary attribute probe AUC (CDL minus CF baseline). Probe AUC
improves substantially in every case, regardless of the direction of the NDCG effect,
confirming that attribute discriminability recovery is a consistent property of CDL
across structural regimes.}
\label{fig:cross_dataset}
\end{figure*}

Attribute discriminability improves consistently: across all four datasets where the CF baseline does not saturate probe AUC, CDL achieves +30--42~pp absolute gains (age on MovieLens-1M +29.6~pp, gender on Last.fm-360K +42.1~pp, stars on Yelp +30.6~pp, age on BookCrossing +30.1~pp). On Last.fm-360K, Yelp, and Audience Factory both NDCG and discriminability improve simultaneously; on MovieLens-1M and BookCrossing, CDL accepts a controlled NDCG cost ($-7.2\%$ to $-15.8\%$) in exchange for substantial attribute gains. Notably, on Last.fm-360K CDL at $\lambda=2.0$ prevents premature early stopping (epoch~13 at $\lambda \leq 1.0$) and trains for 41 epochs, reaching NDCG 0.638 vs.\ 0.592 for the CF baseline --- CE supervision acting as a regulariser against shallow local optima.

\begin{figure*}[t]
\centering
\includegraphics[width=\textwidth]{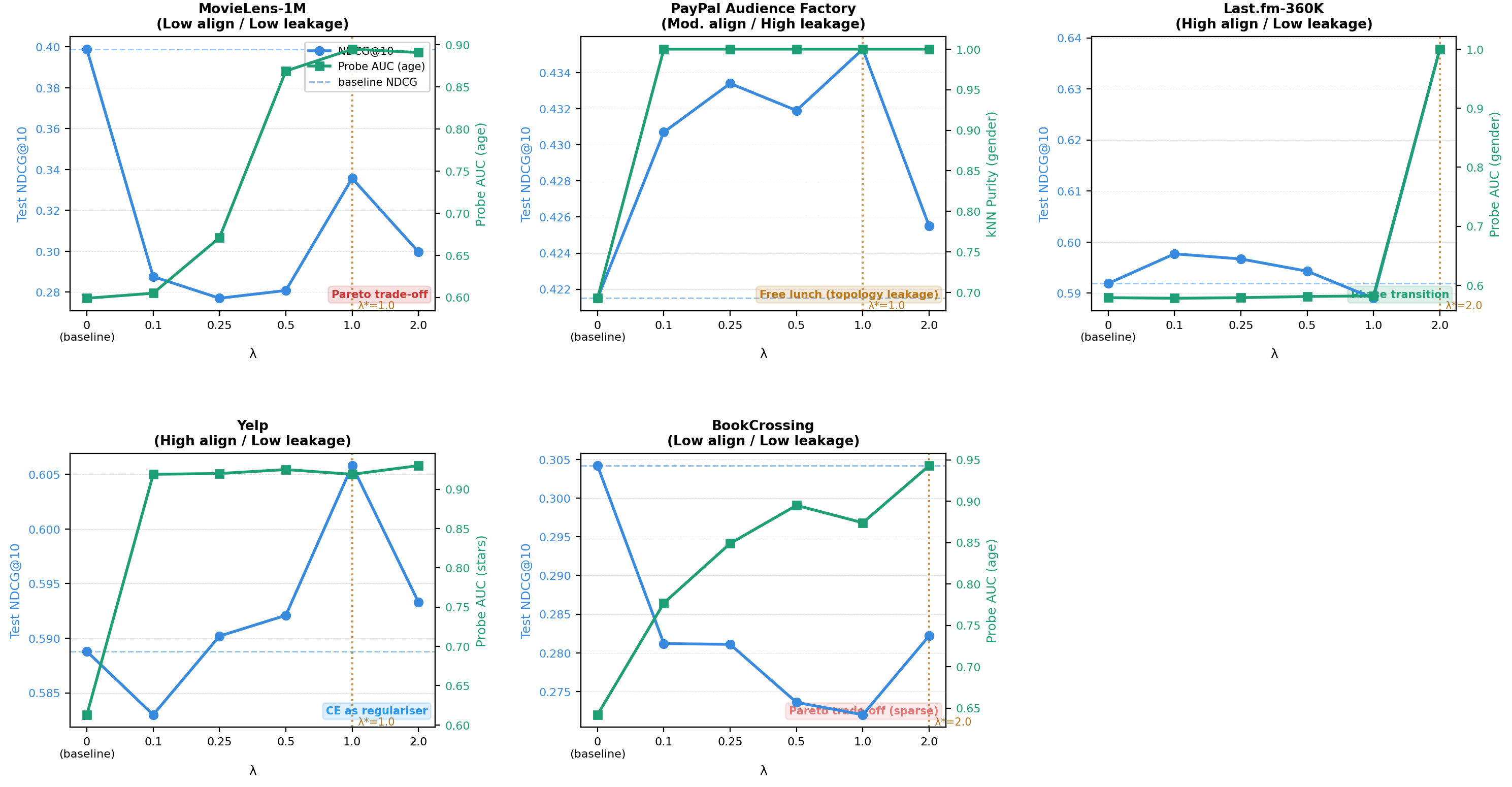}
\caption{\textbf{$\lambda$-sweep Pareto curves across all five datasets.} Each panel
plots test NDCG@10 (blue, left axis) and primary attribute probe AUC (teal, right axis)
as a function of $\lambda$; the filled circle at $\lambda=0$ is the CF baseline
(\texttt{otm\_only}) with a dashed reference line at its NDCG value. The panel label
identifies the dataset's regime on the semantic alignment $\times$ topology leakage axes.
\emph{Pareto trade-off} (MovieLens, BookCrossing): NDCG declines smoothly as probe AUC
rises --- $\lambda$ navigates the frontier. \emph{Phase transition} (Last.fm): both
curves are flat until $\lambda=2.0$, where a simultaneous jump in NDCG and probe AUC
signals a threshold effect. \emph{CE as regulariser} (Yelp): NDCG rises then falls,
with a clear optimum at $\lambda=1.0$. \emph{Free lunch} (Audience Factory): NDCG is
flat and probe AUC is already saturated at all $\lambda$ due to topology leakage.}
\label{fig:lambda}
\end{figure*}

Table~\ref{tab:lambda} and Figure~\ref{fig:lambda} report NDCG@10 and probe AUC across the full $\lambda$ grid for each dataset.

\begin{table*}[t]
\centering
\caption{$\lambda$-sweep results. Best NDCG@10 per dataset is \textbf{bolded}.
$\uparrow\uparrow$ denotes the knee where both NDCG and AUC simultaneously improve most.}
\label{tab:lambda}
\small
\begin{tabular}{lcccccccccc}
\toprule
 & \multicolumn{2}{c}{MovieLens-1M} & \multicolumn{2}{c}{Last.fm-360K} & \multicolumn{2}{c}{Yelp} & \multicolumn{2}{c}{BookCrossing} & \multicolumn{2}{c}{Audience Factory} \\
\cmidrule(lr){2-3}\cmidrule(lr){4-5}\cmidrule(lr){6-7}\cmidrule(lr){8-9}\cmidrule(lr){10-11}
$\lambda$ & NDCG & AUC & NDCG & AUC & NDCG & AUC & NDCG & AUC & NDCG & kNN \\
\midrule
0.10 & 0.288 & 0.832 & 0.598 & 0.670 & 0.583 & 0.757 & 0.281 & 0.777 & 0.431 & 1.000 \\
0.25 & 0.277 & 0.854 & 0.597 & 0.671 & 0.590 & 0.762 & 0.281 & 0.849 & 0.433 & 1.000 \\
0.50 & 0.281 & 0.881 & 0.594 & 0.673 & 0.592 & 0.768 & 0.274 & 0.895 & 0.432 & 1.000 \\
1.00 & \textbf{0.336} & 0.907 & 0.589 & 0.670 & \textbf{0.606}$^{\uparrow\uparrow}$ & 0.766 & 0.272 & 0.874 & \textbf{0.435} & 1.000 \\
2.00 & 0.300 & 0.909 & \textbf{0.638}$^{\uparrow\uparrow}$ & 0.989 & 0.593 & 0.770 & \textbf{0.282} & \textbf{0.943} & 0.426 & 1.000 \\
\bottomrule
\end{tabular}
\end{table*}

\section{Analysis: Semantic Alignment and Topology Leakage}
\label{sec:analysis}

The $\lambda$-sweep across five datasets reveals two qualitatively distinct regimes.
We characterise them by two measurable graph properties that explain the observed
behaviour and allow practitioners to predict which regime they will encounter before
running experiments.

\paragraph{Semantic alignment.}
We define \emph{semantic alignment} as the degree to which the one-to-one attribute
is predictive of user preference, i.e.\ the mutual information between the attribute
label and the set of items a user interacts with. High alignment means that users
with the same attribute tend to prefer the same items; low alignment means the attribute
and preferences are approximately independent.

\begin{itemize}
\item \textbf{High alignment} (Last.fm, Yelp, Audience Factory): Music listening
  preferences on Last.fm are strongly correlated with nationality and age group ---
  knowing a user's country substantially reduces uncertainty about their preferred
  artists. On Yelp, a business's city and star rating directly predict which users
  will review it. On Audience Factory, the behavioural segments encoding user intent
  are partially determined by demographic group membership. In all three cases, CDL
  improves NDCG@10 because CE supervision on attributes provides a signal that is
  \emph{aligned} with the ranking objective.

\item \textbf{Low alignment} (MovieLens, BookCrossing): Age, gender, and occupation
  are weakly correlated with movie and book preferences at the population level. CE
  supervision on these attributes provides a gradient signal that is \emph{orthogonal
  or antithetical} to the ranking objective, and CDL incurs a NDCG cost.
\end{itemize}

\paragraph{Topology leakage.}
We define \emph{topology leakage} as the extent to which message passing on the graph
\emph{already} encodes the one-to-one attribute structure into node embeddings, even
without CE supervision. Leakage occurs when OTO attribute nodes are directly connected
to source nodes in the graph --- every GNN layer propagates attribute information
through the shared encoder regardless of the loss.

The Audience Factory graph includes explicit user--gender and user--age edges used for
message passing. As a result, the CF baseline (\texttt{otm\_only}) already achieves
linear probe AUC of 1.0 for both attributes from early in training. CE supervision
cannot further improve discriminability because it is already perfect through leakage.
CDL's benefit on Audience Factory therefore manifests entirely in the \emph{ranking}
improvement (+3.3\% NDCG@10), not in attribute discriminability.

On MovieLens-1M, the same edges are present, yet probe AUC for age (0.599) and
occupation (0.507) remain far below ceiling under the CF baseline. This is because
21-way occupation classification creates sparse message-passing signal that does not
leak through adequately at the graph's density. CE supervision provides the discriminative
signal that topology cannot.

\paragraph{Two-axis framework.}
Table~\ref{tab:two_axis} places all five datasets on the semantic alignment $\times$
topology leakage grid. Each cell corresponds to a qualitatively distinct regime
in the $\lambda$-sweep of Figure~\ref{fig:lambda}.

\begin{table}[h]
\centering
\caption{\textbf{Dataset regimes on the two-axis framework.}
\emph{Low alignment, low leakage}: CDL recovers attribute discriminability at a
controllable NDCG cost; $\lambda$ navigates the Pareto frontier.
\emph{High alignment, low leakage}: CE supervision reinforces the ranking signal
--- both NDCG and probe AUC improve simultaneously.
\emph{High leakage}: attribute structure is already encoded via message passing;
CDL's benefit is purely in ranking quality.}
\label{tab:two_axis}
\vspace{2pt}
\begin{tabular}{lcc}
\toprule
 & \multicolumn{2}{c}{\textbf{Semantic Alignment}} \\
\cmidrule(lr){2-3}
\textbf{Topology Leakage} & Low & High \\
\midrule
Low  & MovieLens-1M, BookCrossing & Last.fm-360K, Yelp \\
High & ---                        & Audience Factory \\
\bottomrule
\end{tabular}
\end{table}

\paragraph{Practical guidance.}
A practitioner can position a new dataset on this grid before running experiments:
(1) Compute a simple correlation score between attribute labels and item co-occurrence
patterns (e.g.\ average pairwise Jaccard similarity of interaction sets, stratified by
attribute class) to estimate semantic alignment. (2) Inspect whether OTO attribute nodes
are directly adjacent to source nodes in the graph to assess leakage. These two checks
require only the raw graph structure and take seconds to compute.

If semantic alignment is high, set $\lambda = 1.0$ as a default starting point. If
alignment is low but leakage is also low, use the $\lambda$-sweep to find the
operating point on the Pareto frontier. If leakage is high, CDL still improves ranking
quality and embedding purity without requiring attribute discriminability improvement.

\section{Conclusion}
\label{sec:conclusion}

We identified a silent representational failure in heterogeneous GNN training: applying
BPR uniformly to mixed-cardinality graphs causes one-to-one attribute embeddings to
collapse to near-random geometry, invisible to standard ranking metrics. We proposed
Cardinality-Decomposed Loss (CDL), which resolves this by assigning objectives by
structural cardinality (BPR for OTM, CE for OTO), confirmed the underlying gradient
conflict mechanistically, and evaluated the method across five diverse datasets.

CDL consistently recovers attribute discriminability and improves ranking quality when
attributes are semantically aligned with user preferences (+7.8\% NDCG on Last.fm-360K,
+2.9\% on Yelp, +3.3\% on Audience Factory), while enabling a controlled Pareto
trade-off when they are not. The two-axis framework of semantic alignment and topology
leakage provides a principled and computationally cheap way to predict CDL's behaviour
on a new dataset before running full experiments.

\paragraph{Limitations.}
The optimal $\lambda$ is dataset-specific and currently requires a small sweep to
identify. An adaptive $\lambda$ schedule that responds to the gradient cosine
similarity signal during training is a natural extension. Additionally, CDL's
performance on hyperbolic geometry embeddings or contrastively pre-trained backbones
has not been evaluated and may exhibit different trade-off characteristics.
Finally, sharper demographic embeddings have a dual-use character: they support
fairness auditing and transparent user segmentation, but could equally facilitate
more precise demographic targeting in advertising systems.

\section*{Acknowledgements}

We thank Giuseppe Vitis from PayPal for insightful discussions. We also thank Suraj Singireddy for his contributions to this project during his tenure at PayPal. We'd also like to extend thanks to the
PayPal AI organization for their support.

\bibliographystyle{plain}
\bibliography{cdl_references}

\end{document}